\pdfoutput=1

\documentclass[11pt]{article}

\usepackage[final]{acl}

\usepackage{times}
\usepackage{latexsym}

\usepackage[T1]{fontenc}

\usepackage[utf8]{inputenc}

\usepackage{microtype}

\usepackage{inconsolata}
\usepackage{hyperref}

\usepackage{caption}
\usepackage{subcaption}

\usepackage{soul}
\usepackage{graphicx}
\usepackage{multirow}
\usepackage{pifont}
\usepackage{array,booktabs,ragged2e}
\newcolumntype{R}[1]{>{\RaggedLeft\arraybackslash}p{#1}}
\newcolumntype{L}[1]{>{\RaggedRight\arraybackslash}p{#1}}

\usepackage{enumitem}
\usepackage{tcolorbox}
\usepackage{amsmath}
\usepackage{xcolor,colortbl}

\definecolor{valgood}{HTML}{d0e0e3}

\definecolor{valbest}{HTML}{d9ead3}

\usepackage{changepage}

\usepackage{circledsteps}

\newcommand\myCircled[2][]{\ifmmode
\Circled[fill color=black,inner color=white,#1]{\mathsf{#2}}
\else
\Circled[fill color=black,inner color=white,#1]{\sffamily#2}
\fi
}

\newcommand{\methodology}{\textsc{Culture Cartography}}
\newcommand{\tool}{\textsc{Culture Explorer}}

\newcommand*{\imgg}[1]{%
    \raisebox{-.03\baselineskip}{%
        \includegraphics[
        height=0.65\baselineskip,
        width=0.65\baselineskip,
        keepaspectratio,
        ]{#1}%
    }%
}
\newcommand{\delete}{\imgg{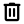}}

\newcommand{\edit}{\imgg{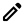}}
\newcommand{\regenerate}{\imgg{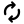}}

\usepackage{ntheorem}
\theoremseparator{:}

\title{Culture Cartography: Mapping the Landscape of Cultural Knowledge}

\newcommand{\treelogo}{\raisebox{5pt}{\includegraphics[scale=0.050]{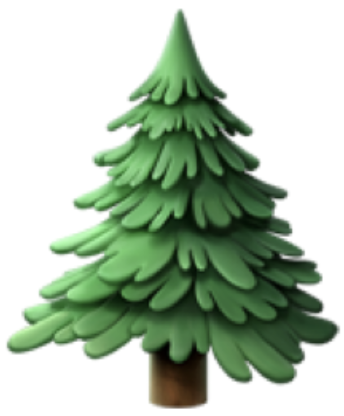}}}
\newcommand{\metalogo}{\raisebox{4pt}{\includegraphics[scale=0.004]{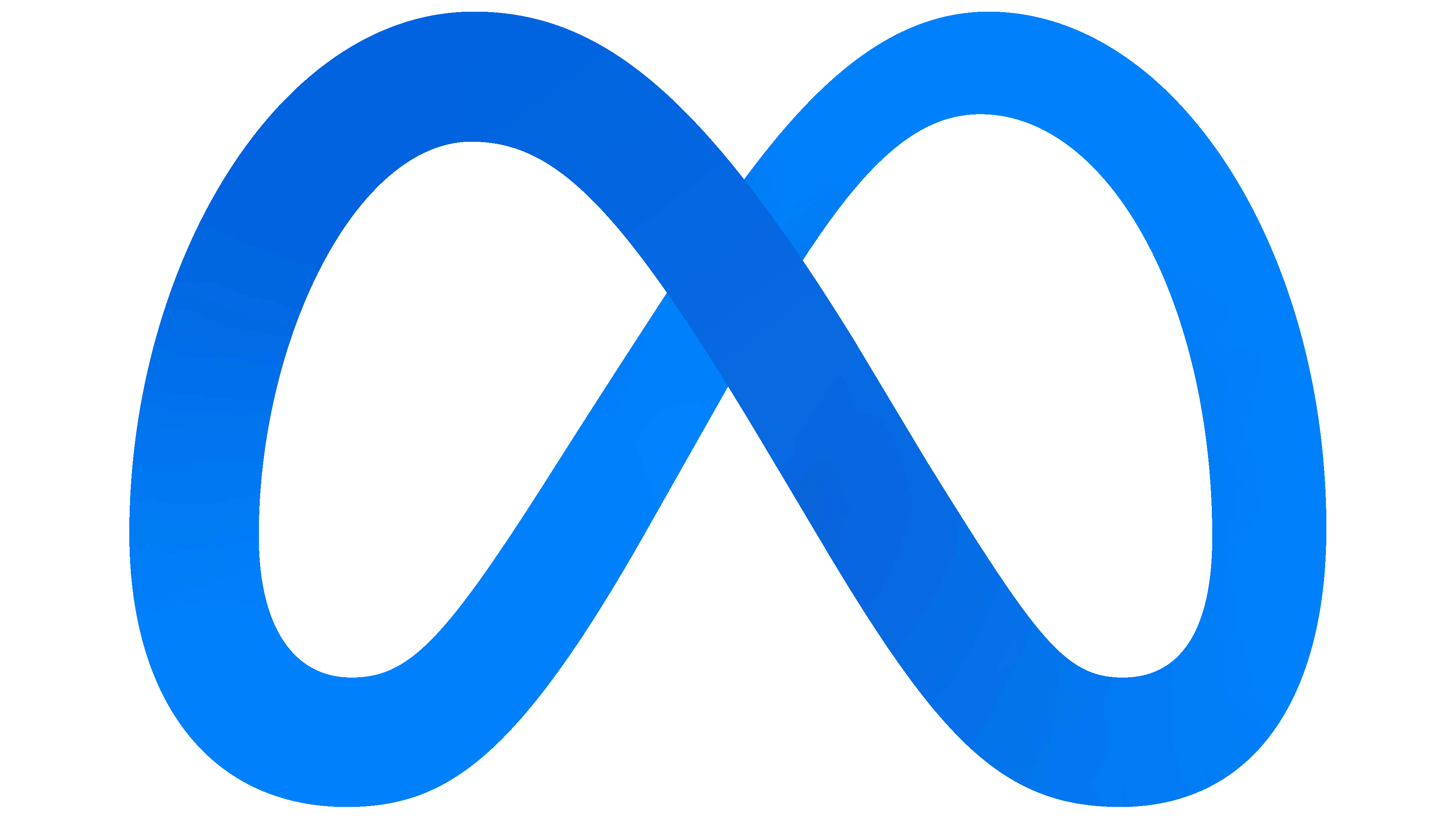}}}
\newcommand{\gtlogo}{\raisebox{3.4pt}{\includegraphics[scale=0.025]{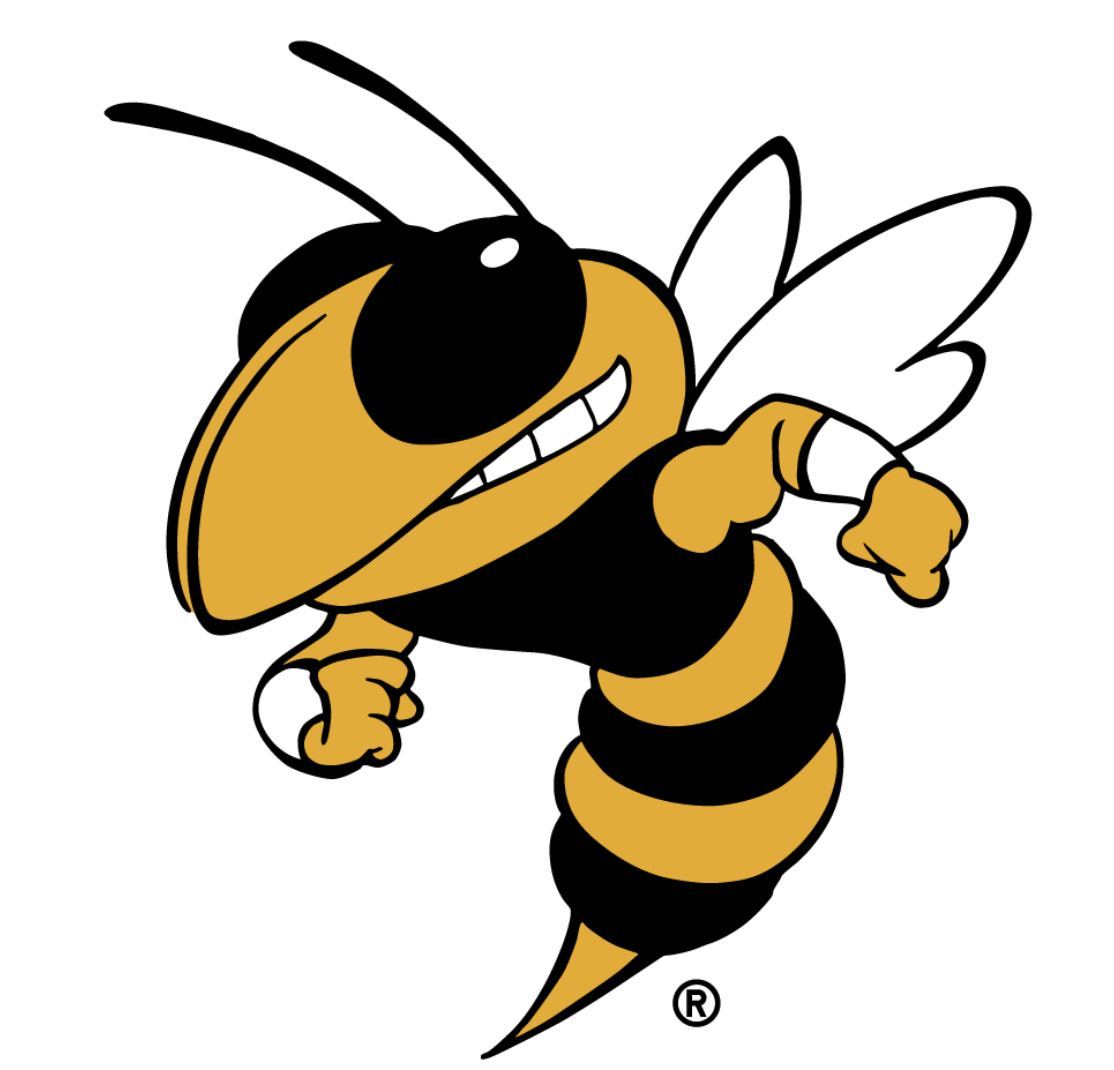}}}

\newcommand{\gt}{\gtlogo}
\newcommand{\stanf}{\treelogo}
\newcommand{\meta}{\metalogo}

\author{Caleb Ziems \stanf \hspace{0.3em}
        William Held \gt \hspace{0.3em}
        Jane Yu \meta \hspace{0.3em}\\
        \textbf{Amir Goldberg} \stanf \hspace{0.3em}
        \textbf{David Grusky} \stanf \hspace{0.3em}
        \textbf{Diyi Yang} \stanf \\
        \stanf Stanford University
        \gt Georgia Institute of Technology \meta Meta AI \\
        \texttt{\small\{cziems, held, amirgo, grusky, diyi\}@stanford.edu}
}

\begin{document}
\maketitle

\begin{abstract} 
To serve global users safely and productively, LLMs need culture-specific knowledge that might not be learned during pre-training. How do we find knowledge that is (1) {salient} to in-group users, but (2) {unknown} to LLMs? The most common solutions are \textit{single-initiative}: either researchers define challenging questions that users passively answer (traditional annotation), or users actively produce data that researchers structure as benchmarks (knowledge extraction). The process would benefit from \textit{mixed-initiative} collaboration, where users guide the process to meaningfully reflect their cultures, and LLMs steer the process to meet the researcher's goals. We propose \methodology{} as a methodology that operationalizes this mixed-initiative vision. Here, an LLM initializes annotation with questions for which it has low-confidence answers, making explicit both its prior knowledge and the gaps therein. This allows a human respondent to fill these gaps and steer the model towards salient topics through direct edits. We implement \methodology{} as a tool called \tool{}. Compared to a baseline where humans answer LLM-proposed questions, we find that \tool{} more effectively produces knowledge that strong models like DeepSeek R1, Llama-4 and GPT-4o are missing, even with web search. Fine-tuning on this data boosts the accuracy of Llama models by up to 19.2\% on related culture benchmarks.
\end{abstract}

\section{Introduction}
\begin{figure}[t!]
    \centering \includegraphics[width=\linewidth]{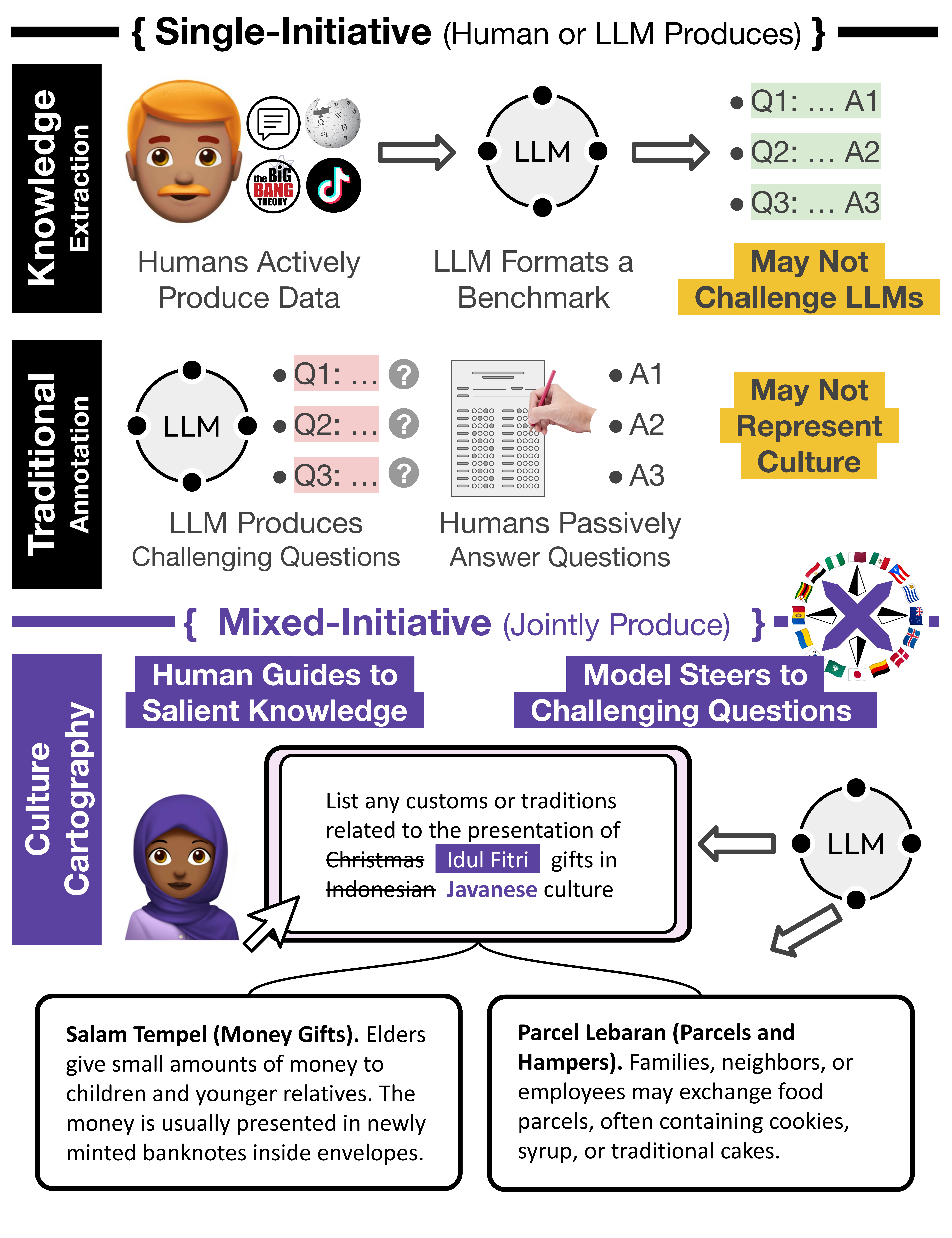}
    \caption{\textbf{\methodology{}} is a new method for identifying culturally-salient knowledge gaps in LLMs. Prior methods are single-initiative: either a human determines the distribution (\textit{Knowledge Extraction}), which may not challenge models, or an LLM decides on challenging questions (\textit{Traditional Annotation}), which may not represent human interests. \methodology{} is the first \textit{mixed-initiative} method that combines four key ingredients: \textbf{(1)} \textbf{an LLM proposes} challenging questions; \textbf{(2)} \textbf{a human proposes} salient questions; \textbf{(3) human edits constrain subsequent LLM generations}; and \textbf{(4)}  \textbf{the data forms a tree} structure. Compared to prior methods, \methodology{} identifies more LLM knowledge gaps.} 
    \label{fig:crown_jewel}
\end{figure}

Large Language Models (LLMs) can empower users to be more knowledgeable, productive, and creative \citep{carmichael2024ipsos,adiguzel2023revolutionizing,yang2024social,chen2021evaluating,si2024can}, but this utility is often diminished for under-represented groups \citep{cao2023assessing,yonglow2024,ziems2023multi} and cultures \citep{myung2024blend,shi-etal-2024-culturebank,chiu2024culturalteaming}, due to data imbalances in pre-training \citep{li2024attributing} and post-training \citep{ryan2024unintended}. LLM agents that \textit{lack} knowledge about their user's cultures can be less helpful personal assistants \citep{qiu2024evaluating}, less relevant recommender systems \citep{casillo2023context}, and less engaging conversation partners \citep{cao-etal-2024-bridging}, while being more prone to generate harmful stereotypes \citep{cheng2023compost} and violate social norms \citep{santy2023nlpositionality}.

\textbf{Problem.} A principal challenge is identifying the cultural knowledge that is both necessary for language models to effectively serve in-group users, and also absent from current models' awareness. The most common methods for finding missing knowledge involve benchmarking with single-initiative datasets (Figure~\ref{fig:crown_jewel}, \textit{top}). These approaches generally follow one of two patterns: either researchers define questions that challenge LLMs~\citep{li2023coannotating} and have human annotators provide answers (identifying hard but potentially non-salient knowledge), or they convert existing cultural knowledge into benchmarks using LLMs (capturing salient but potentially not challenging knowledge). Ideally, users and current-LLMs should be involved in a mixed-initiative interaction~\citep{horvitz1999principles} (Figure~\ref{fig:crown_jewel}, \textit{bottom}) where humans guide the process towards culturally salient knowledge, and LLMs guide the process towards knowledge missing from existing training data.

\textbf{Proposed Solution.} We propose \methodology{} as the first mixed-initiative annotation method to satisfy the above desiderata with all of the following ingredients:
\begin{enumerate}
    \item \textbf{An LLM proposes challenging questions} for which it has low confidence in its answers, thus exposing its knowledge \textit{gaps}---the domain of interest for many researchers.
    \item \textbf{The human makes direct edits or proposes new questions} that reflect their expertise and interests, thus introducing cultural salience.
    \item \textbf{Human edits will guide and constrain subsequent LLM generations}, thus making the interaction truly mixed-initiative.
    \item \textbf{Knowledge is visualized in a tree data structure}, thus affording humans more control through parallel exploration.
\end{enumerate}

Figure~\ref{fig:crown_jewel} (\textit{bottom}) exemplifies each of these ingredients. Here, the LLM proposes a low-confidence question about Christmas gift-giving, which does not align with the annotator's expertise. She is Muslim, like the vast majority of people who live on the Indonesian island of Java, so she edits the question to ask instead about Javanese gift-giving during Eid al-Fitr, an important Islamic holiday. This directly informs the LLM's updated answer suggestions, which are structured as a tree.

\begin{figure*}[t!]
    \centering \includegraphics[width=\linewidth]{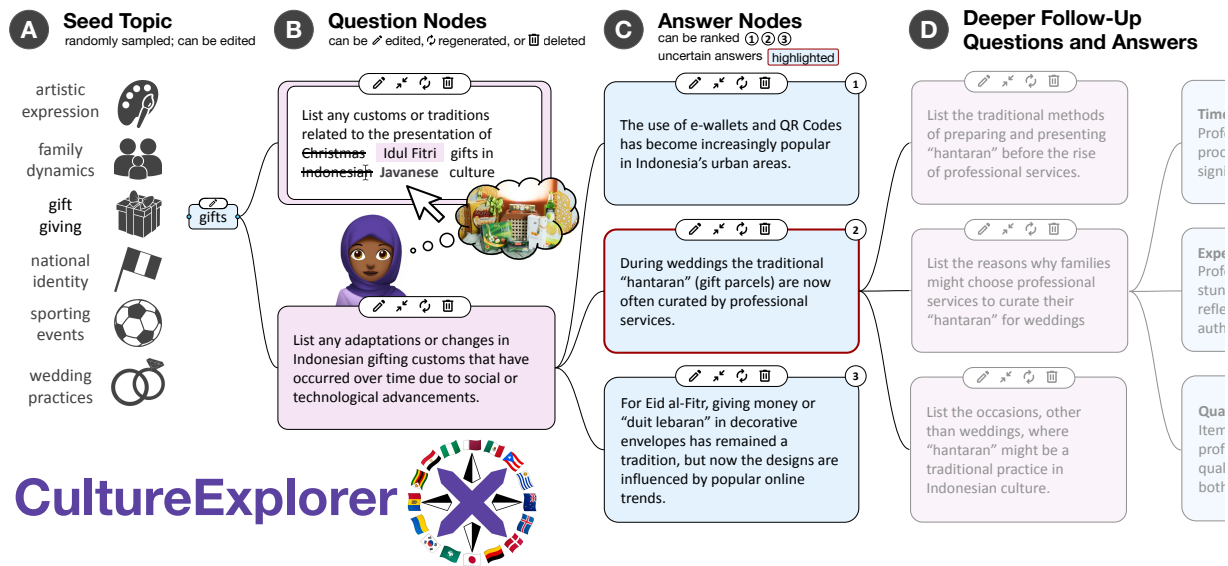}
    \caption{\small{
    \textbf{The \tool{} interface} allows human experts to lead the annotation process, as they can \edit{}~Edit, \regenerate{}~Regenerate, or \delete{}~Delete  nodes at any time.
    Cultural Knowledge annotation is initiated with \myCircled{A} a seed topic (here: \textit{gifts}), which the LLM uses to generate \myCircled{B}~Question nodes. Here, the annotator is editing the first Question node to make it more specific to her Islamic culture. Each Question will serve as a seed for the LLM to generate \myCircled{C}~Answer nodes. The user can then pick the questions and answers interests her, clarify through edits, or write her own from scratch, iteratively expanding the tree with \myCircled{D}~deeper follow-up questions and answers.}
    }
    \label{fig:culture_explorer}
\end{figure*}

\textbf{Research Tool.} Annotation in such a broad and nebulous domain as culture could seem prohibitively inefficient, but we implement a tractable solution with an open-source web-tool called \tool{} (Figure~\ref{fig:culture_explorer}). This tool not only solves the more mundane aspects of the annotation task, like boilerplate and text formatting, but also visually facilitates the mixed-initiative interaction. 
Unlike constrained and linear chat-based interfaces, \tool{} affords users a more interactive interface to consider and edit a growing tree of knowledge. In this way, \tool{} empowers users with a sense of \textit{direct manipulation} \citep{shneiderman1983direct}. The user can make edits rapidly, reversibly, and iteratively, while the tool visually displays the ramifications of these edits by generating parallel follow-up questions. By expanding and pruning branches, the user can consider multiple thematic directions and then focus on what most interests them. 

\textbf{Findings.} We use \tool{} to build cultural knowledge banks for two multicultural and multiethnic countries: \textit{Nigeria} and \textit{Indonesia}. By design, we expect \tool{} will outperform single-initiative methods (Figure~\ref{fig:crown_jewel}) at eliciting knowledge that is both salient and challenging. Indeed, compared to \textit{traditional annotation}, \tool{} identifies data that is at least 6\% less likely to be known by DeepSeek R1, and up to 42\% less likely to be known by other models. Furthermore, unlike \textit{knowledge extraction}, \tool{} produces data that is not easily discoverable online. We find search-enabled models do not outperform search-disabled models at recalling our data. Finally, we demonstrate how our methodology is aligned with the objectives of the field via transfer learning experiments. By fine-tuning on data produced with \tool{}, we can boost the downstream performance of LLMs on other culture benchmarks by up to 19.2\% accuracy.

\textbf{Contributions.} In summary, we propose \methodology{}, a new methodological framework for eliciting culturally-salient knowledge gaps in LLMs. We implement the idea as \tool{} and demonstrate its utility over prior methods. We publicly release all artifacts, including data, code, tooling, and models.\footnote{\small{\texttt{\href{https://github.com/SALT-NLP/culture-cartography}{github.com/SALT-NLP/culture-cartography}}}}

\section{Related Work}
\label{sec:related_work}

There are strong economic, social, and scientific motivations to build more culturally-competent language models that are capable of meaningfully engaging with users from different cultural backgrounds \citep{hershcovich2022challenges}. For common ground, effective language technologies need knowledge of the behavioral norms \citep{shi-etal-2024-culturebank,sky2023sociocultural,rao2024normad}, %
linguistic conventions \citep{shaikh2023modeling}, values \citep{cao2023assessing}, and preferences \citep{kirk2024prism} that shape each user's interactions with the model. This motivates new culture-specific training and evaluation datasets \citep{hershcovich2022challenges}. There are many datasets for advancing cultural competence in specific NLP tasks \citep{shode-etal-2023-nollysenti,muhammad2023afrisenti,muhammad2022naijasenti,tonneau2024naijahate,ilevbare2024ekohate,vargas2024hausahate,olamma2019hidden,adelani2022few,olatunji2023afrispeech,owodunni2024accentfold}, but in the current task-agnostic paradigm, QA-style knowledge benchmarking is the most common \citep{adilazuarda-etal-2024-towards}. Benchmarks are typically built via \textit{Knowledge Extraction} or \textit{Traditional Annotation}.  

\textbf{Knowledge Extraction.} 
To scale up evaluation, LLMs can distill knowledge from web sources \citep{nguyen2024cultural,nguyen2023extracting} like Wikipedia \citep{fung2024massively,li2024well,naous-etal-2024-beer}, TV \citep{fung2023normsage}, or social media \citep{shi-etal-2024-culturebank}. LLMs can also generate synthetic evaluation data without seed knowledge \citep{wang-etal-2024-cdeval,liu2024omgeval}, which human annotators prune and validate. But there are concerns of test set contamination \citep{oren2023proving} when distilled or synthetic data overlaps with LLM pre-training data, and such knowledge extraction is limited only to higher-resource cultures that are well-represented on the web \citep{seth2024dosa}. Alternatively, unstructured \citep{zhang2009unstructured} or semi-structured interviews \citep{karatsareas2022semi} can provide knowledge from under-represented cultures. However, these methods are not designed to target the gaps in models' knowledge.

\textbf{Traditional Annotation.} For the sake of efficiency, it is standard for annotators to respond to pre-determined questions that target gaps in models' knowledge. Questions can derive from secondary sources like templates \citep{yin-etal-2022-geomlama, myung2024blend}, LLM generations \citep{liu2024omgeval,ziems2023normbank}, social media \citep{sky2023sociocultural,huang2023culturally}, or human experts \citep{masala2024vorbe}. Games can serve as a more dynamic and interactive alternative to questionnaires \citep{seth2024dosa,shaikh2023modeling}, but these also depend on fixed seed topics. In each case, the culture informer is not empowered to steer the topical distribution of the data they are providing.

\section{\tool{}}
\subsection{Tool Design}
\label{sec:tool_culture_explorer}
Cultural competence is not a generalizable objective. Like many other forms of alignment, the task is ambiguous \citep{tamkintask2022,li2023eliciting}, as its formalization depends on the culture and the specific users in question. The problem we aim to solve is \textbf{mixed-initiative elicitation}: guiding members of a cultural group to specify what cultural competence means to them, and to do so in an efficient manner. An efficient solution will not waste human effort to reproduce what language models already know, but rather prioritize regions of knowledge that current models lack, following the literature on active learning \citep{cohn1994improving}.

\tool{} is our proposed solution that balances flexibility with efficiency,  empowering users to co-construct with the LLM a branching tree of cultural knowledge. As shown in Figure~\ref{fig:culture_explorer}, this tree is composed of related questions and answers, and users can edit, add, or delete elements at any time. At each iteration, the LLM generates question and answer suggestions. Similar to Generative Active Task Elicitation~\citep{li2023eliciting}, this allows \tool{} to act as a data scaffolding and brainstorming tool that guides respondents towards critical knowledge gaps. But moving beyond a linear chat, \tool{} preserves the respondents' creative freedom to both refine specific answers through edits and define the topic space by adding and removing nodes. 
 
The \tool{} interface is built on \textsc{Farsight}\footnote{\textsc{Farsight} was designed for visualizing AI harms.} \citep{wang2024farsight}, and specially tuned for the relevant culture domains (for details on our specific prompting methodology, see Appendix~\ref{appendix:prompts}). We now explain the pipeline in more detail, in steps corresponding to Figure~\ref{fig:culture_explorer}.

\paragraph{Adding Knowledge.} \tool{} is organized as a tree data structure, with nodes representing branching questions and their answers, which users expand recursively. Annotation is initiated by a pre-selected but editable seed topic\footnote{Seeds derive indirectly from \citet{brown2004human} Human Universal categories. We feed each Brown Universal into \tool{}, and generate for each seed a tree of depth 6 (3 rounds of questions and answers). By semantically clustering the answers across 8 national cultures, we  identify new universals to seed the user-facing version of \tool{}. We cluster across: \textsl{Argentina, Australia, Germany, India, Indonesia, Nigeria, Saudi Arabia,} and \textsl{the United States.}} Figure~\ref{fig:culture_explorer}, \myCircled{A}), which the LLM uses to generate up to 5 Question nodes ( \myCircled{B}). The user can then pick a question that interests her, or clarify through edits, or write her own. Once the user is satisfied with a Question, the AI will generate as many as 5 Answer nodes ( \myCircled{C}). With Uncertainty Estimation \citep{kivlichan2021measuring}, \tool{} visually highlights answer nodes in which the LLM is not confident. \tool{} measures model confidence by prompting the same model, ``\textit{Does this answer the question correctly?}'' It constrains the logits to \textit{True/False}, and takes the probability of \textit{True} as the answer confidence. Answers with confidence below a threshold ($\leq 0.4$) are marked as \textit{uncertain}.

The user can edit, regenerate, or delete any node at any time, and in response, the tool will generate new follow-up questions ( \myCircled{D}), visually displaying the ramifications of the user's changes. Moreover, \tool{} explicitly encourages annotators to make significant edits and novel contributions, as it sums the edit distance over all contributions and computes a scalar monetary reward.

\paragraph{Scoring Answers.} A user may not choose to edit every answer given by the LLM, but users can still provide a valuable preference signal by scoring LLM answers for their relevance and personal applicability. \tool{} asks users to score AI answers on a 0-3 Likert Scale, where 3 awards ``best'' answers that can't be improved, and 0 marks ``bad'' or incorrect answers.

\subsection{Data Collection}
\label{sec:dataset}

\begin{table*}[ht]
\centering
\small
\resizebox{\textwidth}{!}{%
\def\arraystretch{1.15}
\begin{tabular}{lc|ccc|cc|cc}
 \toprule
& \textbf{Ethnolinguistic} & \multicolumn{3}{c}{\underline{\textbf{Synthetic Data}}} & \multicolumn{2}{c}{\underline{\textbf{Traditional Annotation}}} & \multicolumn{2}{c}{\underline{\textbf{\methodology{}}}}\\
\textbf{Country} & \textbf{Groups} & \textbf{Scored Answers} & \textbf{Score ICC} & \textbf{Avg. Score} & \textbf{Fixed Answers} & \textbf{Pref. Pairs} & \textbf{Free Answers} & \textbf{Pref. Pairs}\\
\midrule
Nigeria & 7 & 1,913 & 0.58 & 2.6 / 3 & 944 & 757 & 521 & 262 \\
Indonesia & 13 & 3,412 & 0.55 & 2.5 / 3 & 1,081 & 468 & 586 & 1,196\\
 \bottomrule
\end{tabular}}
\caption{\textbf{\methodology{} Dataset Statistics} demonstrate the size of the data we collected ($\sim$1k answers to fixed questions; $\sim$500 free answers with \tool{}) and its reliability  ($ICC\geq0.55$ on Score annotations, which is moderate), as well as the cultural diversity of our annotator pool (7 distinct ethnolinguistic groups from Nigeria, and 13 distinct groups from Indonesia). Finally, we see that AI responses are quite reliable for these cultures, with quality scores that are better than ``good'' on average (e.g., avg = 2.5/3.0 for Nigeria).\label{tab:dataset_stats}} 
\end{table*}

\begin{figure*}[t!]
    \centering \includegraphics[width=0.98\linewidth]{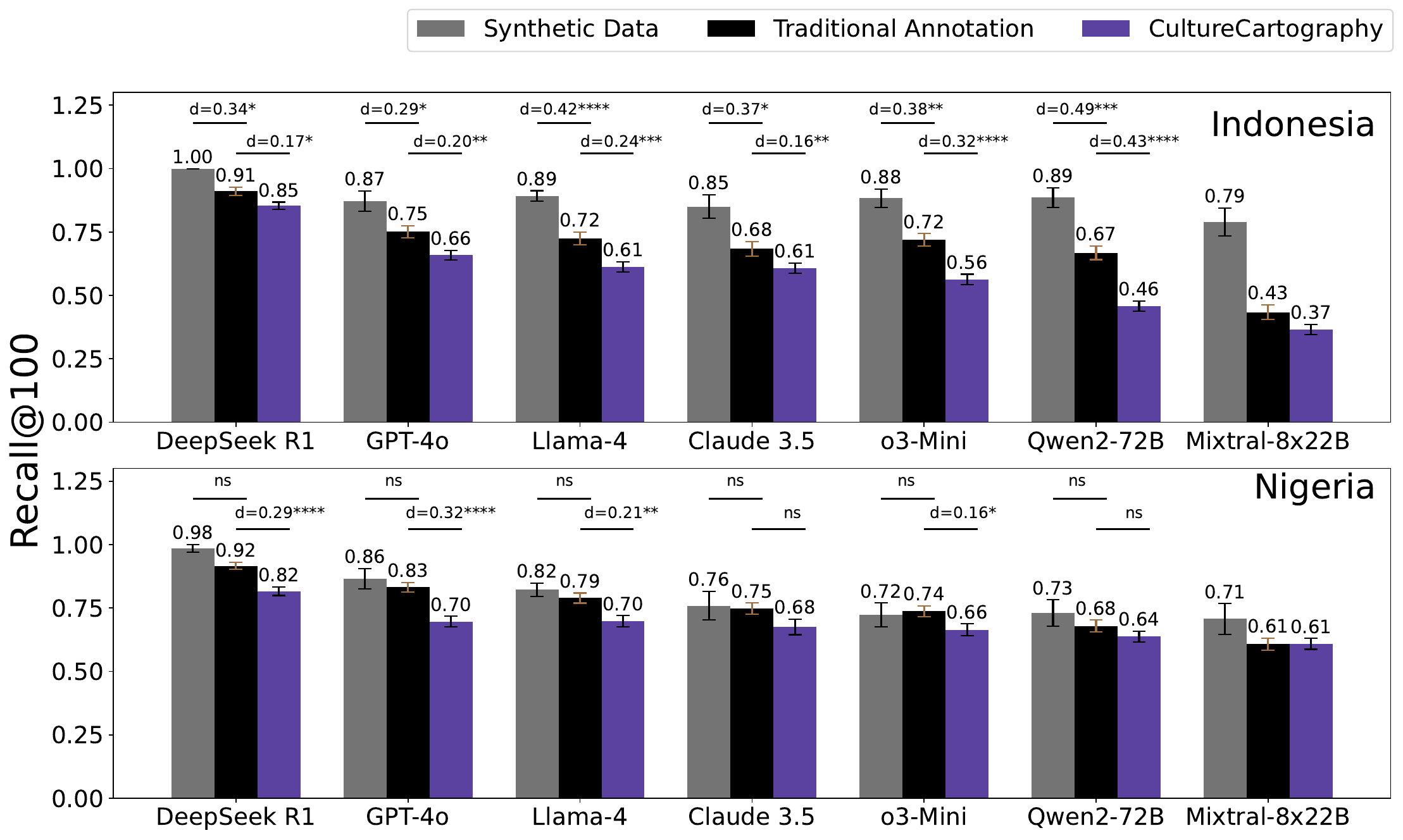}
    \caption{
    \textbf{Performance on \methodology{}.} Powerful models like DeepSeek R1 can entirely solve \textit{Synthetic Data} ($R@100\geq98$\%), and also perform well on \textit{Traditional Annotation} data ($R@100\leq$92\%). Most importantly, \methodology{} data is appreciably harder than these single-initiative data sources, with moderate and statistically significant effect sizes {\small (\texttt{ns} = ``not significant''; $^{*}$~$p < 0.05$; $^{**}$~$p < 0.01$; $^{***}$~$p < 0.001$; $^{****}$~$p < 0.0001$)} for both R1 {\small($d=0.17$ Indonesia; $d=0.29$ Nigeria)} and GPT-4o {\small ($d=0.20$ Indonesia; $d=0.32$ Nigeria)}.}
    \label{fig:eval_cartography}
\end{figure*}

\tool{} allows us to fill many of the gaps identified in Related Work (\S\ref{sec:related_work}). We can build a participatory, multilingual knowledge bank of localized cultural knowledge that complements what LLMs already know. We focus our data collection on two culturally diverse yet under-resourced countries: \textit{Nigeria} and \textit{Indonesia}. Each nation contains hundreds of distinct ethnolinguistic groups, with over 500 distinct indigenous languages in Nigeria \citep{campbell2008ethnologue}, and over 600 ethnic groups across Indonesia \citep{ananta2015demography}.

The following was approved by the Institutional Review Board at the authors' institution. We recruit annotators on Upwork, aiming for balance across ethnolinguistic groups (see Appendix~\ref{appdx:annotator_demographics} for details). Annotations are collected in the national language for each respective country, since this language is shared across ethnolinguistic groups (\textit{Bahasa Indonesia} for Indonesian annotators, and \textit{English} for Nigerian annotators). To establish baselines for the \methodology{} methodology and test our principal hypothesis that it is better than entirely model-driven approaches, we collect data in three distinct, non-overlapping subsets:

\noindent \textbf{(1) Synthetic Data:} Humans validate the top-four answers given by the LLM to a pre-determined set of \textit{fixed questions} about \textit{behavioral norms}.\footnote{We created this set of questions by running \tool{} to a tree depth of 6 and filtering for \textit{behavioral norms} with a FastText classifier that we distilled from GPT-4. For other aspects of data collection with \tool{}, the underlying LLM was \texttt{gpt-3.5-turbo}. Thus we show the benefits of the \methodology{} approach do not depend on the most recent advances in language modeling.} For each question, annotators scored the AI answers for quality on the same 0-3 Likert Scale. We retained only high-quality AI answers: for a given question, we identified all pairs of AI answers whose quality scores were different with statistical significance by t-test ($\alpha = 0.05$) and kept only the better answer from each pair.

\paragraph{(2) Traditional Annotation:} With the same fixed questions about behavioral norms, the respondents, having considered the top-four \textit{AI Responses} above, then provided up to four new answers that complemented what the AI already gave. Annotators were explicitly encouraged to think of specific examples from their most specific and local cultures about what AI wouldn't already know. In this way, the \textit{Traditional Annotation} directly mirrored the incentive structure of \methodology{} to identify gaps in LLM knowledge. Importantly, in this subset, respondents could not edit questions or guide their topical distribution.

\textbf{(3) \methodology{}:} This represents the set of all answers that humans wrote from scratch, working with the \tool{} in an unconstrained manner, where they could edit AI questions freely and iteratively. Annotators worked on the task for as long as they liked, and were paid a fair hourly rate, plus bonuses for their total edit distance on questions and answers.

\subsection{Dataset Summary}
\label{subsec:dataset_summary}
Table~\ref{tab:dataset_stats} gives the summary statistics for the three annotated subsets of \methodology{}. We worked with 19 Indonesian annotators from across 13 ethnolinguistic groups and 12 provinces, as well as 9
Nigerian annotators from 7 ethnolinguistic groups across 5 states (see Tables \ref{tab:annotator_demographics_nga} and \ref{tab:annotator_demographics_ind} in Appendix~\ref{appdx:annotator_demographics} for more details). From each pool of Nigerian and Indonesian annotators, we collected $\sim$1k fixed answers with \textit{Traditional Annotation}, and $\sim$500 free answers with \methodology{}. Annotators also scored $>5$k LLM answers. These score annotations are reliable, with a moderately high inter-annotator agreement of $ICC\geq0.55$, which is a moderate intraclass correlation \citep{shrout1979intraclass}, and reasonable for the subjective nature of this task.

On the left side of Table~\ref{tab:dataset_stats}, we see that the LLM's responses are quite reliable, with quality scores that are better than ``good'' on average (e.g., avg = 2.5/3.0 for Nigeria). For all three countries, over 90\% of answers were deemed at least passable (avg. score $\geq$ 1.0) --- less than 10\% of AI answers were deemed fully incorrect. One can interpret these results like precision metrics, suggesting that, when a well-prompted LLM answers its own pre-determined cultural questions about Nigerian and Indonesian cultures, the answers are reliable.

\section{Evaluating \methodology{}}
\label{sec:evaluation}
In this section, we test whether leading LLMs recall less \methodology{} data than they recall synthetic or traditionally annotated data. We select seven flagship models to evaluate. There are three proprietary API models: GPT-4o \citep{hurst2024gpt}, o3-Mini\footnote{We use the \textit{medium} reasoning setting for o3-Mini.} \citep{o3mini}, and Claude 3.5 Sonnet \citep{anthropic2024claude}. The remaining four models are open-weight: DeepSeek R1 \citep{guo2025deepseek}, Llama-4-Maverick \citep{meta2025llama4}, Qwen 2-72B \citep{yang2024qwen2technicalreport}, and Mixtral-8x22B. To scalably evaluate model awareness of gold answers, we rely on LLM-as-a-Judge evaluations with GPT-4o to compute the Recall@$K$

{\small\begin{equation*}
R@K = \frac{|\{\text{gold answers}\} \cap \{\text{model answers @$K$}\}|}{|\{\text{gold answers}\}|}
\end{equation*}}

Here, \{model answers @$K$\} refers to the set of all $K$ answers produced by the model by iteratively prompting it: ``\textit{We're looking different examples. Without explanation, list 10 more examples.}''\footnote{This parallels the instructions given to humans to find examples that \textit{AI wouldn't already know}.} The LLM-as-a-Judge determines the overlap in the numerator by iterating through each gold answer and telling us: ``\textit{Does any part of \{model answers @$K$\} contain the same information as the \{gold answer\}?} '' by answering ``Yes'' or ``No.'' For all subsequent experiments, we set $K=100$ because baseline model performance on \textit{Synthetic Data} plateaus here (see Figure~\ref{fig:recall_k} in Appendix~\ref{appdx:validating_llm_as_judge}).

Human validation also ensures that our LLM-as-a-Judge approach is reliable. One author blindly annotated a random sample of data, oversampling for the minority class. We observed a substantial agreement of 85\% between human and model judgments, with a Cohen's $\kappa=0.66$, which indicates substantial agreement \citep{mchugh2012interrater}.  

\paragraph{Q1: Is \methodology{} Data More Challenging?}
\label{subsec:evaluation_explorer}
Figure~\ref{fig:eval_cartography} compares \methodology{} data (in purple) against both baselines: traditional annotation (in black), and synthetic data (in gray). We see that, compared to traditional annotation, Indonesian \methodology{} data is 6\% less likely to be known by R1 (0.85 vs. 0.91), and Nigerian data is 10\% less likely to be known by R1 (0.82 vs. 0.92). \methodology{} is even less likely to be known by other models. This difference between \methodology{} and Traditional Annotation is statistically significant by t-test $(\alpha = 0.05)$ on both Indonesian and Nigerian data for each of the following models: DeepSeek R1, GPT-4o, Llama-4, and o3-Mini. The effect sizes on R1 are Cohen's $d=$0.17 and $d=$0.29 for Indonesian and Nigerian performance gaps, indicating small or moderately-sized effects. We conclude that \methodology{} more readily produces challenging and long-tail cultural knowledge than does synthetic or traditional annotation.

\paragraph{Q2: How do LLMs Compare on \methodology{}?}
Figure~\ref{fig:eval_cartography} also demonstrates that many LLMs cannot reach high levels of recall for \methodology{} data, thus demonstrating gaps in their long-tail knowledge of pluralistic cultures. DeepSeek R1 completely saturates the \textit{Synthetic} subset, maintains a relatively high recall on \textit{Traditional Annotation} data (91\% Indonesia; 92\% Nigeria), and achieves the highest overall performance on \methodology{} (85\% Indonesia; 82\% Nigeria). 
In contrast to R1, Mixtral-8x22B fails to produce as much as half of the \methodology{}. Most models are between these two extremes and attain moderate scores, with recall around 60-70\% on the \methodology{} subsets. When we sort models by their performance on \methodology{}, we get the same relative order for both countries: DeepSeek R1 $\succ$ GPT-4o $\succ$ Llama-4 $\succ$ Claude 3.5 $\succ$ o3-Mini $\succ$ Qwen2-72B $\succ$ Mixtral-8x22B. There is not a stark strong performance gap between API-based and open-weight models here, as both the best and the worst performing model are open-weight, and the runner-up model is the proprietary GPT-4o. Furthermore, reasoning models do not unanimously win: o3-mini ($\sim$200B parameters) falls behind Claude 3.5 Sonnet ($\sim$175B), a slightly smaller, non-reasoning model. To conclude, \methodology{} allows produces stable evaluation results that reveal nuanced differences in model performance, not attributable to reasoning or model size alone.

\paragraph{Q3: What Are the Knowledge Gaps?}
Even the best reasoning model, DeepSeek R1, fails to recall 15-18\% of \methodology{} data. Now we investigate the topical distribution of this missing knowledge. We do so by adapting LlooM \citep{lam2024concept}, a concept induction algorithm. First, for each QA pair that DeepSeek R1 fails to recall, we prompt GPT-4o to summarize the QA pair with 3 bullet points of at most 30 words each. Next we use $k$-means clustering over the full set of bullet point sentence embeddings \citep{reimers2019sentence} to produce the top 10 semantic clusters each for the Nigerian and Indonesian subsets respectively. Then we prompt GPT-4o to perform the LlooM \texttt{Synthesize} operator and summarize each bullet point cluster with two key concept patterns each. A concept pattern consists of a text label and a corresponding prompt for classification. Finally, we use the concept prompts to classify the knowledge that DeepSeek R1 originally missed. 

Table~\ref{tab:lloom_topics} lists DeepSeek R1's top 5 most prevalent missing concepts for Nigerian and Indonesian cultures respectively. DeepSeek R1's knowledge gaps here are not merely incidental trivia; they concern topics that are essential for preserving the social cohesion of families and larger communities. We see \textit{Community Engagement} appears in almost 80\% of missing Nigerian knowledge, and almost a third of missing Indonesian knowledge. For example, R1 was unaware of the Bornean communal meal called \textit{baseprah} in which people of different social status dine together, fostering the spirit of \textit{gotong royong} or communal responsibility and unity.

\begin{table}[t!]
\small
\resizebox{\columnwidth}{!}{%
\def\arraystretch{1.15}
\begin{tabular}{lr}\toprule 
\multicolumn{2}{c}{Nigerian \methodology{} Data}\\\midrule
\textbf{Concept} & \textbf{Proportion} \\ 
1. Community Engagement & 79.2\% \\
2. Cultural Preservation & 77.1\% \\
3. Family Roles & 30.2\% \\
4. Funeral Rituals & 9.4\% \\
5. Family and Community Integration & 8.3\% \\ \midrule
\multicolumn{2}{c}{Indonesian \methodology{} Data}\\\midrule
\textbf{Concept} & \textbf{Proportion} \\ 
1. Cultural Adaptations & 48.8\% \\
2. Exclusive Cultural Practices & 44.2\% \\
3. Cultural Traditions & 38.4\% \\
4. Cultural Gatherings & 38.4\% \\
5. Community Engagement & 31.4\% \\
\bottomrule 
\end{tabular}
}
\caption{\textbf{DeepSeek R1's top 5 most prevalent missing concepts} for the Nigerian and Indonesian subsets of \methodology{} data. Note that categories are not mutually exclusive, so proportions add to more than 100\%.} 
\label{tab:lloom_topics}
\end{table}

\paragraph{Q4: Is \methodology{} Google-Proof?}
\label{subsec:evaluation_gpt_4_5}
If \methodology{} can produce data not found on the web, we can demonstrate another benefit of our methodology over knowledge extraction methods, or other benchmarks collected from the internet, which are prone to test set contamination \citep{oren2023proving}. We ask, is \methodology{} \textit{Google-Proof}'' \citep{rein2024gpqa}? That is, could the challenging questions from \methodology{} be answered by a flagship LLM with retrieval access to the web. For direct comparison with our Figure~\ref{fig:eval_cartography} results, we evaluate GPT-4o with web search enabled,\footnote{a.k.a., \texttt{gpt-4o-search-preview}} and compare these results to GPT-4o's prior performance without search. We further estimate performance for the most advanced frontier model currently available in Appendix~\ref{appdx:gpt_4_5}.

Results in Table~\ref{tab:search_results} show that web search fails to improve the performance of GPT4o on \methodology{}. In fact, performance is worse with search (54.8\% with, vs. 69.7\% recall without search on \texttt{nga}; $p<0.0001$). The exact mechanism behind the performance drop is speculative, but it is conceivable that web search narrows the model's focus to the head of the knowledge distribution, since the model recalls less of the long tail. 

The principal conclusion we can draw is that \methodology{} is Google-Proof. Knowledge derived from this methodology is not easily retrieved from public web sources, including those used for Knowledge Extraction in \S\ref{sec:related_work}, like Wikipedia, television transcripts, and social media. This complements our previous findings, further demonstrating the benefits of \methodology{} over single-initiative data collection methods. Such methods would be more easily solved by web search, as we will show in \S\ref{sec:results_transfer}.
  
\begin{table}[t!]
\small
\resizebox{\columnwidth}{!}{%
\def\arraystretch{1.15}
\begin{tabular}{lcc}\toprule 
\multicolumn{3}{c}{GPT4o $R@100$ on \methodology{}}\\\midrule
\textbf{Culture} & \texttt{ind} & \texttt{nga} \\ 
\cmidrule{1-3}
{no search} &  \textbf{65.9} & \textbf{69.7}\\
\textit{with search} &  {61.9} & {54.8}\\ \midrule
Effect Size ($d$) & 0.08 & 0.31\\
Significance & ns & {\tiny $p<0.0001$}\\
\bottomrule 
\end{tabular}
}
\caption{\textbf{\methodology{} is Google-Proof.} GPT4o attains lower Recall@100 scores on \methodology{} with web search enabled, vs. without search. Since performance does not improve with search, we conclude that \methodology{} is Google-Proof.} 
\label{tab:search_results}
\end{table}

\section{Transfer Performance}
\label{sec:results_transfer}

Section \ref{sec:evaluation} showed that, compared to synthetic or traditional annotations, \methodology{} data is more challenging, and unlike knowledge extraction, \methodology{} helps circumvent test set contamination, as it appears \textit{Google-Proof}. Now we demonstrate that data produced with \methodology{} quantifiably aligns with the objectives of prior efforts in culturally-aware NLP, since training on \methodology{} data can boost the downstream transfer performance of LLMs on related culture benchmarks. 

For completeness, we evaluate transfer performance on benchmarks that represent contrasting annotation paradigms. First, we consider BLEnD \citep{myung2024blend}, which represents a more traditional annotation approach where annotators responded to fixed questions from a set of 500 pre-defined question templates, resulting in the largest available benchmark for Indonesian and Nigerian cultural knowledge. Next, we look at {CulturalBench} \citep{chiu2024culturalteaming}, which was produced with a red-teaming methodology similar to our mixed-initiative approach, where humans and LLMs jointly produce knowledge. In CulturalBench, humans propose social situations, and LLMs generate related MCQ questions; finally humans modify the questions until they can stump the LLM. The key difference between CulturalBench and \methodology{} is that CulturalBench was produced in through a linear chat interaction rather than a tree-based exploration, and humans iteration did not impact the topical domain of LLM generations in real time.

\paragraph{Training.} Given compute limitations, we opted to train two relatively smaller models, Llama-3.1-8B and Qwen-2-7B, on answers and preference pairs from either \methodology{} or our \textit{Traditionally Annotated} data from \S\ref{sec:dataset}. Training has two steps. First, we derive a set of high-quality preference pairs. A preference pair is given for every set of AI answers whose scores are different with statistical significance by t-test with $\alpha=0.05$. Following the method of \citet{shaikhaligning2025}, we treat Human Responses as strictly preferable to any AI Responses. Then we train with SFT on the preferred answers, followed optionally by DPO on the derived preference pairs (see Appendix~\ref{appendix:hyperparameters} for hyperparameters). 

\textbf{Results.} Table~\ref{tab:downstream_tasks} shows the transfer performance on the Nigerian (\texttt{nga}) and Indonesian (\texttt{ind}) subsets of BLeND and CulturalBench. We begin our discussion with the Llama results. Only \methodology{} data (\textit{Cart.}) helps Llama models significantly outperform vanilla models on both benchmarks. After SFT+DPO on \methodology{}, Llama-3.1-8B achieves $+6.5\%$ accuracy on BLEnD-\texttt{nga}, and $+7.1\%$ accuracy on BLEnD-\texttt{ind} compared to vanilla ($p<0.0001$ by paired t-test). \methodology{} also significantly boosts vanilla models by $+18.2\%$ and $+19.2\%$ on CulturalBench-\texttt{nga} and CulturalBench-\texttt{ind} respectively ($p<0.05$). This demonstrates how \methodology{} produces data that aligns with prior benchmarking efforts.

Training on \methodology{} data also results in overall better downstream performance than training on the \textit{Traditionally Annotated} data (\textit{Trad.}, Table~\ref{tab:downstream_tasks}). For example, on BLEnD-\texttt{nga}, performance is $+3.2\%$ better with \textit{Cart.} than \textit{Trad} ($p<0.0001$). While these benefits are similar on CulturalBench, the results do not reach statistical significance with such a small test set (26 Indonesian QA pairs in CulturalBench, vs. 18.5k pairs in BLEnD). This further the added utility of \methodology{} for collecting knowledge that more richly reflects underlying cultures. 

The findings are directionally the same for the Qwen2-7B: training on \methodology{} data results in better downstream performance than training on \textsl{Traditionally Annotated} data. With SFT+DPO on data from \methodology{}, Qwen2-7B achieves +3.9\% accuracy CulturalBench-\texttt{ind}, +4.5\% accuracy on CulturalBench-\texttt{nga}, and +0.6\% accuracy on BLEnD-\texttt{nga} compared to the vanilla model. These boosts are not as large, nor statistically significant as those observed with Llama-3.1-8B, but this is expected since vanilla Qwen-2 starts with higher baseline performance, more than 10\% greater than vanilla Llama-3.1-8B on these benchmarks.  

\begin{table}[t]
\centering
\small
\resizebox{\columnwidth}{!}{%
\def\arraystretch{0.8}
\begin{tabular}{clcccccccc}
 \toprule
& &  \multicolumn{2}{c}{\textbf{BLEnD}} & \hspace{0.5pt} & \multicolumn{2}{c}{\textbf{CulturalBench}}\\
\cmidrule{3-4} \cmidrule{6-7}
&& \textbf{\texttt{nga}} & \textbf{\texttt{ind}} && \textbf{\texttt{nga}} & \textbf{\texttt{ind}} \\
 \midrule \midrule
Llama-3.1-8B & Vanilla& 56.0& 66.5&& 50.0& 46.2&\\ \midrule
\parbox[t]{4mm}{\multirow{4}{*}{\rotatebox[origin=c]{90}{\textit{Trad.}}}} \\
& SFT& 55.8& 68.7&& \textbf{59.1}& \textbf{53.8}&\\
& SFT+DPO& 59.3& \textbf{73.2}$^{***}$&& {50.0}& \textbf{65.4}&\\ 
& \\ \midrule
\parbox[t]{4mm}{\multirow{4}{*}{\rotatebox[origin=c]{90}{\textit{Cart.}}}}\\
& SFT& 57.9& 69.1&& \textbf{68.2}$^{*}$ & \textbf{61.5}&\\
& SFT+DPO& \textbf{62.5}$^{***}$& \textbf{73.6}$^{***}$&& \textbf{63.6}& \textbf{65.4}$^{*}$ &\\ 
& \\ \midrule \midrule
Qwen2-7B & Vanilla& 64.9 & 79.8 && 68.2 & 69.2 &\\ \midrule
\parbox[t]{4mm}{\multirow{4}{*}{\rotatebox[origin=c]{90}{\textit{Trad.}}}} \\
& SFT& 63.7 & 79.6 && 63.6 & 65.4 &\\
& SFT+DPO& 63.6 & 77.4 && 68.2 & 65.4 &\\ 
& \\ \midrule
\parbox[t]{4mm}{\multirow{4}{*}{\rotatebox[origin=c]{90}{\textit{Cart.}}}}\\
& SFT& 63.8 & 79.4  && 63.6 & 65.4 &\\
& SFT+DPO& 65.5 & 78.7 && 72.7 & 73.1 &\\ 
& \\ \midrule \midrule
GPT4o & + search & 76.8 & 91.3 && 95.5 & 84.6\\ \midrule
\textit{Num.}& \textit{evals} & 16.4k & 18.5k && 22 & 26\\
 \bottomrule
\end{tabular}}
\caption{\label{tab:downstream_tasks} \textbf{Zero-shot Cultural Awareness of Llama-3.1-8B and Qwen-2-
7B} after training (SFT) and optionally preference-tuning (+DPO) on either \textit{Traditional Annotation} (\textit{Trad.}) or \methodology{} (\textit{Cart.}), and evaluated on BLEnD and {CulturalBench}, with statistically significant best performances among Llama models \textbf{bolded} {\small ($^{*}$ $p < 0.05$; $^{***}$ $p < 0.0001$)}. \methodology{} results in better downstream performance than \textit{Traditional Annotation} data. Only \methodology{} data results in significantly better performance than vanilla Llama models on \textit{both} evaluation sets. Qwen results are directionally the same, with better downstream performance from \methodology{}, but results are not statistically significant.
}
\end{table}

Furthermore, we see that fine-tuning with \methodology{} helps close the performance gap between much smaller 8B open models and the much larger, proprietary GPT-4o model with search enabled (bottom of Table~\ref{tab:downstream_tasks}). In doing so, \methodology{} lends itself to solutions for building more culturally-aware NLP systems. 

Finally, these results exemplify the comparative advantage \methodology{} has over single-initiative data collection. While web search failed to improve performance on the more challenging \methodology{} data, search nearly saturates both BLEnD-\texttt{nga} and CulturalBench-\texttt{ind}, with accuracies above 90\%. The same search-enabled model achieved only 54.8\% and 61.9\% recall respectively on the Nigerian and Indonesian subsets of our mixed-initiative \methodology{} data (Table~\ref{tab:search_results}).

\section{Conclusion}
Towards the development of culturally-competent language models, we contribute \tool{} (\S\ref{sec:tool_culture_explorer}) as an interactive annotation tool that implements our mixed-initiative \methodology{} methodology.
Compared to single-initiative annotation, \methodology{} better satisfies two motivating desiderata: the data it generates is \textit{more challenging} for models, while also being \textit{more representative} of human interests. Six flagship LLMs attain $\leq$70\% recall on \methodology{} data, even when search is enabled, so we conclude that \methodology{} is Google-Proof, and thus less prone to test set contamination than single-initiative methodologies. Finally, we see that \methodology{} aligns with prior efforts in culturally-aware NLP, since fine-tuning on this data boosts the downstream transfer performance of LLMs on prior benchmarks, and helps close the performance gap between larger and smaller models. To conclude, \methodology{} is a new mixed-initiative method for eliciting useful and representative cultural knowledge. This may complement social science methodologies like surveys and semi-structured interviews, offering a new lens for studying cultural variation and heterogeneity.

\section{Limitations}

\paragraph{Biases in Annotator Recruitment.} 
It is difficult to recruit contributors from under-represented cultures. As a result, the majority of published knowledge banks are exclusively in English \citep{adilazuarda-etal-2024-towards}, and cover the culture groups most accessible on Crowdwork platforms. In this work, we demonstrated a preliminary effort to construct multilingual knowledge resources for low-resourced cultures, but this effort is far from complete.

All of our annotator recruitment was performed through Upwork, which may introduce biases in annotator recruitment, including but not limited to imbalances in underlying annotator distribution on the platform, the ranking algorithm \citep{suhr2021does} or the system of worker reputations on which it is based \citep{thebault2017simulation}. Our recruitment strategy in \S\ref{sec:dataset} was intended to ameliorate some of these factors, as we took a roughly balanced stratified random sample across ethnolinguistic groups for each country we studied. Still, we were limited by time and the availability of annotators, so the number of respondents from each group is relatively small, and as a result, the data we collected may not be fully representative of each group. More importantly, our study is limited to only the responses of those who have stable and reliable access to Upwork and the ability to communicate in English. This is very likely to introduce biases in the worker pool (e.g., by education level, socioeconomic status, etc.).

\paragraph{Culture is More Than Knowledge.} This work builds on existing efforts in Cultural NLP, which focus primarily on benchmarking LLMs to identify critical \textit{knowledge gaps} \citep{adilazuarda-etal-2024-towards}. Our current \tool{} implementation  effectively generates a tree of cultural knowledge, which is useful for benchmarking LLMs. However, culture can be much broader than the domain of factual knowledge \citep{zhou2025culture}, and there are other compelling applications for  \methodology{} more broadly which may be restricted in the current fact-based  implementation of \tool{}. For example, social scientists may be interested in extending \tool{} to help build digital museums that preserve not only knowledge, but also stories, history, and cultural artifacts \citep{srinivasan2005fluid}. To support a more \textit{fluid ontology} in this setting would require further engineering \tool{} beyond its question-answer tree format, and to consider broader themes or abstractions: not only the highly-detailed features of daily life.

\section{Ethics}
\paragraph{Responsible Research Ethics.} This study has been approved by the Institutional Review Board (IRB) at the researchers' institution, and participant consent was obtained using the standard institutional consent form. Annotators were also encouraged to stop any time. They were paid a fair stipend of \$20 per hour for their time. To protect annotators' privacy, all data was anonymized.

\paragraph{Risks in Deployment.} Here we outline risks in deploying \tool{}. Since this tool is powered by a Large Language Model, it shares the risks of many other human-LLM interactions, which include the potential harms of offensive, stereotypical, or hateful outputs, and the risk of misinformation. These risks are mitigated by our task-specific prompts, which constrain the output distribution, and by our use of safety-aligned LLMs. 

Specifically, in this domain, LLMs may misinterpret, flatten, or misrepresent nuanced cultural knowledge. If users do not carefully consider these risks, or overly rely on LLM suggestions, their work will be more prone to endorse dominant cultural narratives at the expense of authentic expression. To mitigate this risk, we explicitly state in the instructions and onboarding video: ``\textit{Please be critical! You know your culture better than AI.}'' For more details on the onboarding, see Appendix~\ref{appdx:onboarding_video_transcript}.

\section*{Acknowledgments}
We are thankful to the members of SALT Lab and the Stanford NLP Group: particularly Camille Harris, Giuseppe Russo, Raj Sanjay Shah, Myra Cheng, Jared Moore, and Sunny Yu for their helpful feedback on the draft. We also immensely appreciated regular feedback from Jing Huang, Julia Kruk, and Yanzhe Zhang in the earliest stages of this project. Caleb Ziems was supported by the NSF Graduate Research Fellowship under Grant No. DGE-2039655. The work was supported in part by a grant from Meta, and by a grant from the Stanford Institute for Human-Centered Artificial Intelligence (HAI).

\bibliography{custom}

\appendix

\section{Annotator Recruitment and Demographics}
\label{appdx:annotator_demographics}
We recruited all annotators on Upwork, offering \$20 per hour, and specifying that applicants needed to be adult workers (18+) from \textit{Nigeria} or \textit{Indonesia} who grew up in the local cultures. To ensure high quality work, we hired only workers with at least 90\% Job Success as indicated by the platform. We also filtered by the following skills: (1) \textit{data annotation}, (2) \textit{data entry}, (3) \textit{writing}, (4) \textit{cultural and ethnic studies}, (5) \textit{arts and culture}. Workers were hired following a roughly balanced stratified random sample across geography (provinces or states), as indicated by the locations in their Upwork profiles (see Tables~\ref{tab:annotator_demographics_nga} and \ref{tab:annotator_demographics_ind} for the results). Then annotators were extensively onboarded by one of the authors of this study. The annotator first watched a 4-minute Instructional Demonstration Video on YouTube, explaining how the \tool{} tool works, as well as the goals of the study. The worker then completed a series of 1-5 staging rounds of annotation to ensure their understanding of the task. Each staging round entailed 15 minutes of work with the \tool{}. The recruiting author evaluated this work and gave extensive personalized feedback, answering questions, and particularly encouraging workers to prioritize novel contributions around knowledge that AI doesn't already know, and to focus on their most local or regional cultures. Once the annotator demonstrated understanding, work was scaled up to 5 hour blocks with random audits for quality. The same pool of annotators contributed to all three data subsets of \S\ref{sec:dataset}: Synthetic Data, Traditional Annotation, and \methodology{}.


\begin{table}
\centering
\resizebox{\columnwidth}{!}{%
\begin{tabular}{lll}
\hline
\textbf{state}     & \textbf{town}     & \textbf{ethnicity} \\ \hline
Akwa   Ibom                 & Uyo           & Ekid               \\ \hline
Anambra State               & Awka          & Igbo               \\ \hline
Federal Capital   Territory & Abuja         & Abawa              \\ \hline
\multirow{2}{*}{Lagos State}                 & Lagos         & Yoruba             \\
               & Lagos         & Yoruba             \\ \hline
\multirow{4}{*}{Ogun State}                  & Abeokuta      &  Egba  \\
                  & Abeokuta      & Ijebu              \\
                  & Abeokuta      & Remo               \\
                   & Ibadan        & Yoruba \\ \bottomrule       
\end{tabular}}
\caption{\textbf{Nigerian Annotator Demographics} for 9 annotators from 7 ethnolinguistic groups across 5 states.
}
\label{tab:annotator_demographics_nga}
\end{table}

\begin{table}
\centering
\resizebox{\columnwidth}{!}{%
\begin{tabular}{lll}
\hline
\textbf{province}     & \textbf{town}     & \textbf{ethnicity} \\ \hline
\multirow{2}{*}{Aceh}         & Banda Aceh        & Acehnese           \\
 & Banda Aceh        & Acehnese           \\ \hline
\multirow{2}{*}{Bali}         & Denpasar          & Balinese           \\
   & Ubud              & Balinese           \\ \hline
Banten       & Tangerang Selatan & Bantenese          \\ \hline
Central Java & Salatiga          & Javanese           \\
East Java             & Malang            &  Javanese        \\ \hline
\multirow{2}{*}{East Kalimantan}        & Balikpapan        & Malay              \\
 & Samarinda         & Malay              \\ \hline
South Kalimantan      & Martapura         & Banjarese          \\ \hline
\multirow{2}{*}{South Sulawesi}        & Makassar          & Bugis              \\
& Makassar          & Bugis              \\ \hline
\multirow{4}{*}{West Java}             & Bandung           & Sundanese          \\ 
            & Indramayu         & Indramayu          \\
             & Jakarta           & Javanese           \\
             & South Tangerang   & Javanese           \\ \hline
West Kalimantan       & Pontianak         & Dayak              \\ \hline
West Nusa Tenggara    & Mataram           & Sasak              \\ \hline
West Sumatra          & Padang            & Minangkabau  \\\bottomrule    
\end{tabular}}
\caption{\textbf{Indonesian Annotator Demographics} for 19 annotators from 13 ethnolinguistic groups across 12 provinces.
}
\label{tab:annotator_demographics_ind}
\end{table}

\section{Validating LLM-as-a-Judge}
\label{appdx:validating_llm_as_judge}
We validate the LLM-as-a-Judge by uniformly sampling 50 evaluations of Nigerian \methodology{} data from across all 7 models, plus 25 additional examples oversampled from the minority-predicted class (here: ``No''). One author blindly annotated these 75 datapoints. The author agreed with the model's judgment 39+25=64 times (85\%), with a substantial Cohen's $\kappa = 0.66$.
\begin{figure*}[t!]
    \centering 
    \begin{subfigure}{.5\textwidth}
      \centering
      \includegraphics[width=\textwidth]{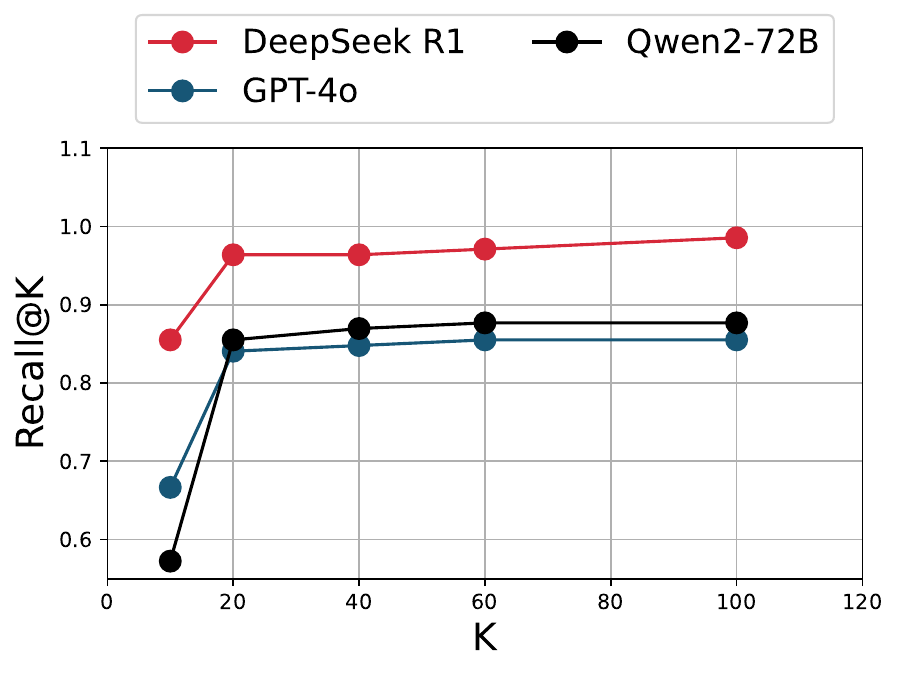}
      \caption{On Synthetic Indonesian Data}
      \label{fig:sub1}
    \end{subfigure}%
    \begin{subfigure}{.5\textwidth}
      \centering
      \includegraphics[width=\textwidth]{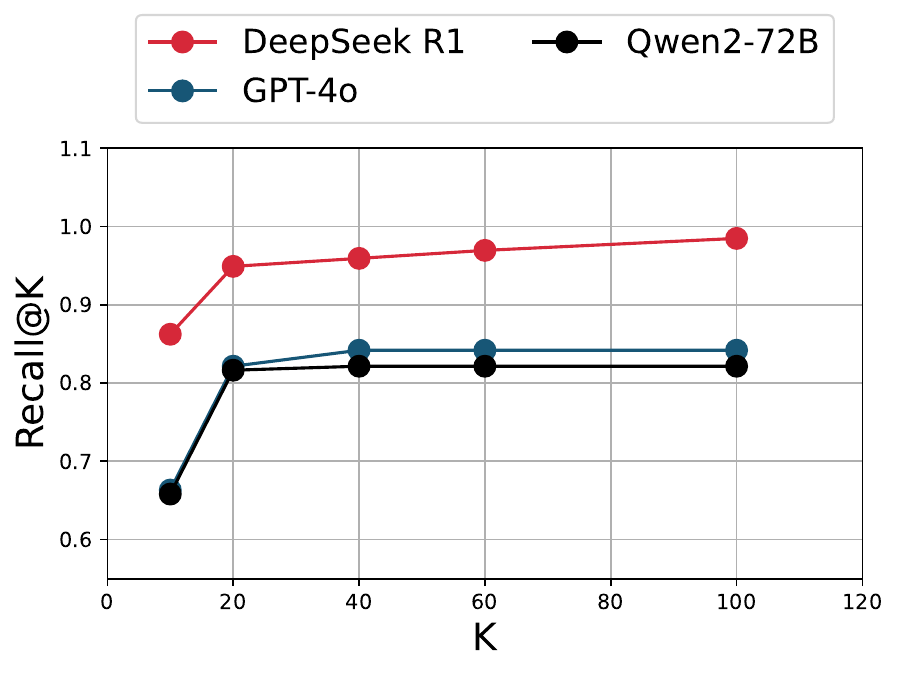}
      \caption{On Synthetic Nigerian Data}
      \label{fig:sub2}
    \end{subfigure}
    \caption{\small{
    \textbf{Recall@K curves for DeepSeek R1, GPT-4o, and Qwen2-72B on \textit{Synthetic Data}} demonstrate that model performances either plateau or reach 100\% by $K=100$}
    }
    \label{fig:recall_k}
\end{figure*}

\section{Estimating GPT-4.5's Performance on \methodology{}}
\label{appdx:gpt_4_5}
At the time of this study, the largest available model was GPT-4.5 \citep{openai2025gpt4_5}. GPT-4.5 would be prohibitively expensive to evaluate on the entirety of \methodology{}. If we use GPT-4o's total evaluation I/O of 66M tokens, then at \$150 / 1M tokens, a full GPT-4.5 evaluation would cost \$9,900. Instead, we estimate an optimistic upper bound on its performance. Given its demonstrated performance benefits over GPT-4o \citep{openai2025gpt4_5}, and our interest in the optimistic upper-bound, we assume GPT-4.5 would correctly answer any question that GPT-4o could answer. We evaluated GPT-4.5 directly on a random sample of 50 questions that GPT-4o got wrong; then we used GPT-4.5's accuracy on this set to interpolate between GPT-4o's performance and perfect performance. We randomly sample 50 questions for which GPT-4o failed to retrieve at least one human gold answer: 25 questions for Indonesia and 25 questions for Nigeria. Using the ChatGPT web interface, one researcher manually evaluated GPT-4.5 on all 50 pairs in a manner that most optimistically estimated upper-bound performance. For each question, the researcher added clarifying details to the prompt that most effectively narrowed the retrieval space without explicitly giving the answer. For example, the answer to one question on ``\textit{coming of age rituals}'' was Otu-Odu, an initiation specific to Igbo women, so the researcher clarified: ``\textit{We're looking different examples of initiation rituals for women in Igbo culture.}'' Then the researcher manually judged the results in the manner of \S\ref{sec:evaluation}. 

In doing so, we computed $R(4.5)_{\sigma_\text{failure}}@K$ --- the recall of GPT-4.5 on the failure set $\sigma_\text{failure}$ --- as 72\% and 48\% for Indonesia and Nigeria respectively. Then we estimated $\widehat{R(4.5)}_{\sigma_\text{full}}@K$, the recall of GPT-4.5 on the full \tool{} set as

{\small\begin{align*}
\widehat{R(4.5)}_{\sigma_\text{full}@K} &= R(4o)_{\sigma_\text{full}}@K \\
& \phantom{==}+ (1-R(4o)_{\sigma_\text{full}}@K)(R(4.5)_{\sigma_\text{failure}}@K)
\end{align*}}

Here we assumed (as an upper bound) that GPT-4.5 would successfully recall anything that GPT-4o recalled, then estimated 4.5's improvement on 4o's failure set according to $R(4.5)_{\sigma_\text{failure}}@K$. Table~\ref{tab:gpt_4_5_results} shows the GPT-4.5 upper bound performance estimates of 90\% recall for Indonesia and 84\% for Nigeria. We see that GPT-4.5 can exceed DeepSeek R1's performance (as given in Figure~\ref{fig:eval_cartography}), but GPT-4.5 does not fully solve these challenging subsets. On \methodology{}, GPT-4.5 lags behind even R1's performance on \textit{Traditional Annotations} (91\% and 92\%), which further demonstrates the ``Google-Proof-ness'' of our approach (Q3). This also further reinforces our \S\ref{subsec:evaluation_explorer} conclusion that \methodology{} produces more challenging data than synthetic or traditional annotation.

\begin{table}[t!]
\small
\resizebox{\columnwidth}{!}{%
\def\arraystretch{1.15}
\begin{tabular}{lcc}\toprule
\textbf{Culture} & \textbf{R@100} & \textbf{95\% CI} \\ \midrule
ind &  0.90 & [0.84, 0.96]\\ 
nga &  0.84 & [0.79, 0.90]\\ 
\bottomrule 
\end{tabular}
}
\caption{\textbf{Expected Performance of GPT-4.5} on the \methodology{} still lag behind best model performances on Fixed Questionnaires, further demonstrating the utility of \tool{} co-annotation for producing challenging and long-tail cultural knowledge.}
\label{tab:gpt_4_5_results}
\end{table}

\section{Prompts}
\label{appendix:prompts}

We build \tool{} on the Farsight \citep{wang2024farsight} codebase, which is licenced under a Creative Commons Attribution International 4.0 License, which permits us to transform and build upon the material for any purpose. Among other changes, we transform the prompts to fit our domain, and these prompts are given below.

\textbf{Nigerian Questions:} \\
\texttt{\small You are an anthropologist who is good at asking important questions about Nigerian culture. Given a description of an abstract cultural concept, please brainstorm 5 specific questions about this concept in Nigerian culture. Put each question in the XML tags: <question></question>. Each question should be written in a way that starts with the word 'List'.\newline cultural concept:Gifts\newline examples: <question>List any customs or traditions related to the preparation and presentation of gifts in Nigerian culture.</question>\newline<question>List the etiquette and expectations surrounding gift-giving and receiving in Nigerian culture.</question>\newline <question>List the differences in gifting practices between various regions or social groups within Nigerian culture.</question>\newline<question>List the occasions when gifts are traditionally exchanged in Nigerian culture.</question>\newline<question>List any adaptations or changes in Nigerian gifting customs that have occurred over time due to social or technological advancements.</question>\newline\newline cultural concept:\{\{concept\}\}\newline examples:}\\

\textbf{Indonesian Questions:} \\
\texttt{\small Anda seorang antropolog yang pandai mengajukan pertanyaan penting tentang budaya Indonesia. Dengan deskripsi konsep budaya abstrak, silakan buat 5 pertanyaan spesifik tentang konsep ini dalam budaya Indonesia. Masukkan setiap pertanyaan dalam tag XML: <question></question>. Setiap pertanyaan harus ditulis dengan cara yang dimulai dengan kata 'Buat'. \newline konsep budaya:Hadiah\newline contoh: <question>Buat daftar kebiasaan atau tradisi yang terkait dengan persiapan dan pemberian hadiah dalam budaya Indonesia.</question>\newline <question>Buat daftar etiket dan harapan seputar pemberian dan penerimaan hadiah dalam budaya Indonesia.</question>\newline <question>Buat daftar perbedaan dalam praktik pemberian hadiah antara berbagai daerah atau kelompok sosial dalam budaya Indonesia.</question>\newline <question>Buat daftar kesempatan saat hadiah secara tradisional dipertukarkan dalam budaya Indonesia.</question>\newline <question>Buat daftar adaptasi atau perubahan dalam kebiasaan pemberian hadiah Indonesia yang telah terjadi dari waktu ke waktu karena kemajuan sosial atau teknologi.</question>\newline\newline konsep budaya:\{\{concept\}\}\newline contoh:} \\

\textbf{Nigerian Answers:} \\
\texttt{\small You are an observant Nigerian person who is good at recalling diverse and accurate traditions, practices, and norms in your culture. Given a question, please brainstorm 5 more specific answers from Nigerian culture. Put each answer in one of the XML tags: <universal> if the answer applies to many cultures, or <local> if the answer applies only to a few related cultures, or <unique> if the answer applies only to Nigerian culture.\newline cultural question: List some Nigerian wedding traditions and what they signify.\newline examples:<unique>Alaga: A Nigerian wedding ceremony officiant whose job is to heckle the groom and his friends as entertainment for the wedding guests. This keeps guests engaged during the hour-long ceremony.</unique>\newline <unique>Aso-Ebi: Nigerian couples choose what their guests wear by assigning a color for the bride’s family and a separate color for the groom’s family.</unique>\newline <local>No Guest List: Nigerian weddings won’t have a guest list. The entire community is welcome and an abundance of food and drink is available in case you end up with your entire community in attendance.</local>\newline <local>Gele: Nigerian brides wear a traditional, ornate headpiece called a Gele. The bridesmaids and families also follow suit and wear a Gele to honor the cultural traditions of the day.</local>\newline\newline cultural concept: \{\{concept\}\}\newline examples:}\\

\textbf{Indonesian Answers:} \\
\texttt{\small Anda seorang antropolog yang pandai mengajukan pertanyaan penting tentang budaya Indonesia. Dengan deskripsi konsep budaya abstrak, silakan buat 5 pertanyaan spesifik tentang konsep ini dalam budaya Indonesia. Masukkan setiap pertanyaan dalam tag XML: <question></question>. Setiap pertanyaan harus ditulis dengan cara yang dimulai dengan kata 'Buat'. \newline konsep budaya: Hadiah\newline contoh: <question>Buat daftar kebiasaan atau tradisi yang terkait dengan persiapan dan pemberian hadiah dalam budaya Indonesia.</question>\newline <question>Buat daftar etiket dan harapan seputar pemberian dan penerimaan hadiah dalam budaya Indonesia.</question>\newline <question>Buat daftar perbedaan dalam praktik pemberian hadiah antara berbagai daerah atau kelompok sosial dalam budaya Indonesia.</question>\newline<question>Buat daftar kesempatan saat hadiah secara tradisional dipertukarkan dalam budaya Indonesia.</question>\newline<question>Buat daftar adaptasi atau perubahan dalam kebiasaan pemberian hadiah Indonesia yang telah terjadi dari waktu ke waktu karena kemajuan sosial atau teknologi.</question>\newline<question>Buat daftar norma budaya untuk mengakui dan menanggapi hadiah dalam budaya Indonesia.</question>\newline<question>Buat daftar signifikansi ukuran, nilai, dan pembungkusan hadiah dalam budaya Indonesia.</question>\newline<question>Buat daftar makna atau simbolisme khusus yang terkait dengan jenis hadiah tertentu di Indonesia budaya.</question>\newline<question>Buat daftar peran hadiah dalam membangun hubungan dan mengekspresikan emosi dalam budaya Indonesia.</question>\newline<question>Buat daftar berbagai jenis hadiah yang umum diberikan dalam budaya Indonesia.</question>\newline\newline konsep budaya:{{concept}}\newline contoh:}\\

\section{Models \& Hyperparameters}
\label{appendix:hyperparameters}
We fine-tune the base version of Llama-3.1-8B on NVIDIA RTX 6000 Ada GPUs with LoRA \citep{hulora2021} (rank 8; $\alpha=16$; dropout of 0.1) for 4 epochs of SFT, followed optionally by 4 epochs of DPO using the Huggingface Trainer and the TRL library \citep{vonwerra2022trl}. We set a batch size of 1, a learning rate of $2e-4$, and an AdamW-8bit optimizer.

\section{Instructions Given to Participants}
\subsection{Recruitment Information}
\textbf{Description:} You are invited to participate in a research study whose goal is to help people partner with AI to collect diverse, culture-specific knowledge, that is organized in a way that makes sense to you. You will also be asked to provide your own knowledge, answering our open-ended questions by listing examples from your local culture. You will also be asked to check the correctness of existing data that has been generated automatically using Generative AI, and if these are wrong, you will have the ability to edit, delete, or regenerate knowledge. 
\textbf{Eligibility:} you must be an adult volunteer (18+) from the country we have specified.
\textbf{Payments:} You will receive \$20 USD per hour as payment for your participation.
\textbf{Risks:} The are no significant risks associated with this study. Study data will be stored securely, minimizing the risk of confidentiality breach. Your individual privacy will be maintained during the research and in all published and written data resulting from the study.

\subsection{Onboarding Video Transcript}
\label{appdx:onboarding_video_transcript}
Welcome to \tool{}, a tool for helping you to visualize and share your own cultural-specific knowledge. You can work with AI to fill in the missing pieces in a growing tree of knowledge. 

When you’re ready, click Yes to start. To get you thinking, the tool starts with a random seed topic, like here is Weddings. Feel free to change this seed to anything that interests you!

According to the instructions, we need to first ask at least 3 interesting questions about our culture. The tool will automatically generate 5 ideas for you, but please be critical! You know your culture better than AI. An interesting question should specifically distinguish your culture from others. The question here about planning customs isn’t very specific, so let’s see what we can do from the instructions: we can either edit, regenerate, add our own, or delete questions. Let’s click regenerate here until we find a better one. This one looks pretty good. The question about etiquette is great because it leads to practical advice that would help an outsider fit in, so we’ll keep that. Finally, an interesting question should be on a topic that is cherished or important to community members. Americans care about money, but none of these questions ask about finances, so we will add one by double clicking the bottom node and typing directly into the box. <types: List some budgetary considerations for American weddings.> You see this follows instructions, since questions should start with the word ``List.'' In the end, we need 3 validated questions. We mark interesting questions by clicking the validation button in the bottom right. 
Notice how the footer keeps track of how many we have. Once we have all three validated, we are told to hover over them and click add answer suggestions. After that, we can click Next on the instructions. 

AI starts us off with 5 answers, which we can again edit or regenerate. Some of these, like dress appropriately, are too vague. Instead of deleting them outright, we can specify: \textit{do not dress in casual attire at a formal ceremony.} Some like do not wear white may be generalized to: \textit{do not wear any clothing that upstages the bride or groom.} Note that each time we make edits, the counter in the right footer updates the bonuses we’ve earned! Bonuses are based on character-level contributions.

The goal is to contribute as much knowledge as you can that is also distinct from what is already in the tree – distinct from what AI already knows. You will work for a fixed amount of time, and you will be paid bonuses for any new knowledge you contribute before the timer runs out. Try to be efficient about finding and contributing new knowledge!

Maybe you have knowledge about regional differences in gift giving, so you select the middle question to expand, revealing AI-generated answers in blue. Notice that some of these answers have a bright teal bar on the right side. These are answers that the AI is confident in. Others have a dark red bar on the right side. These are answers that the AI was NOT confident in, so they will often be wrong. Since your goal is to contribute new knowledge, it is a good idea to start with the dark red bars. Read these highlighted answers and correct them if they are wrong. For example, this last answer about refusing gifts once or twice isn’t true for everyone – it applies only for Chinese Indonesians. You can add:

\begin{quote}
    For Chinese Indonesians, it is polite to refuse a gift once or twice before accepting it, to show modesty and humility. 
\end{quote}

and delete \textit{which is a common practice across various cultures in Indonesia}. At the bottom, you should add any answers you can that aren’t already given by the AI. Here, you might add: 

\begin{quote}
    For Indian or Hindu Indonesians, wrapping should be in bright colors, and leather products and alcohol should not be given. 
\end{quote}

You can also add: 
\begin{quote}
    For Malays and Muslim Indonesians, gifts that have alcohol or pork in them should not be given.
\end{quote}

Let’s say you’ve shared all your knowledge about these topics, and doubt you can contribute more here. Feel free to click ``Hide its answers'' to tidy up your workspace, then move on to a different question.

To find an interesting question that AI doesn’t understand, it may help to go deeper into the tree.  Even if the AI is confident about the initial question on the branch, we can more challenging questions by going deeper! Each time we generate an answer to a question, we can generate more questions about that very answer like this. 

<expands top question, top answer, top question>

Look how this pink question is highlighted red! The red highlighted answer before meant the AI wasn’t confident about that single answer, but a red highlighted QUESTION means it was a very challenging question overall for the AI. Look how all of its answers are red.

This is a great place to contribute your knowledge, especially because an outsider who isn’t familiar with Indonesian culture would have a very hard time learning about this concept of Dutarikh. I tried Googling some terms and found very little documentation. 

If you have other interesting questions to ask, please contribute them here in the ``What Else?'' section. By answering your own questions, you can contribute even more unique and interesting knowledge.

An interesting question should specifically distinguish your culture from others, so you might ask: 

\begin{quote}
    How does Gayo Dutarikh differ from the embroidery of other regions of Indonesia?     
\end{quote}

Then we can use AI to suggest some answers for this question. Note how these answers are quite vague because this topic is challenging for AI. You should definitely add better answers here, or edit existing answers to make them more specific!

Keep building the tree until the timer is up. Finally, click Export to download the tree as a file, which we will upload to Upwork, thus completing this round of annotation.

\end{document}